\documentclass[letterpaper, 10 pt, conference]{ieeeconf}  %

\IEEEoverridecommandlockouts                              %

\usepackage{graphics} %
\usepackage{epsfig} %
\usepackage{mathptmx} %
\usepackage{times} %
\usepackage{amsmath} %
\usepackage{amssymb}  %
\usepackage{xcolor}
\usepackage{graphicx}
\usepackage{tikz} %

\usepackage{algorithm}
\usepackage{algpseudocode}
\usepackage{amsmath} %
\usepackage{arydshln} %
\usepackage[hidelinks]{hyperref}

\title{\LARGE \bf Notes-to-Self: Scratchpad Augmented VLAs for Memory Dependent Manipulation Tasks}

\author{Sanjay Haresh$^{1}$, Daniel Dijkman$^{1}$, Apratim Bhattacharyya$^{1}$, Roland Memisevic$^{1}$%
\\ \texttt{\url{https://qualcomm-ai-research.github.io/Notes-to-Self/}}
\thanks{$^{1}$Qualcomm AI Research}%
\thanks{*Qualcomm AI Research is an initiative of Qualcomm Technologies, Inc.}%
}

\begin{document}

\maketitle
\thispagestyle{empty}
\pagestyle{empty}

\begin{figure*}[thpb]
  \centering
  \includegraphics[width=0.95\textwidth]{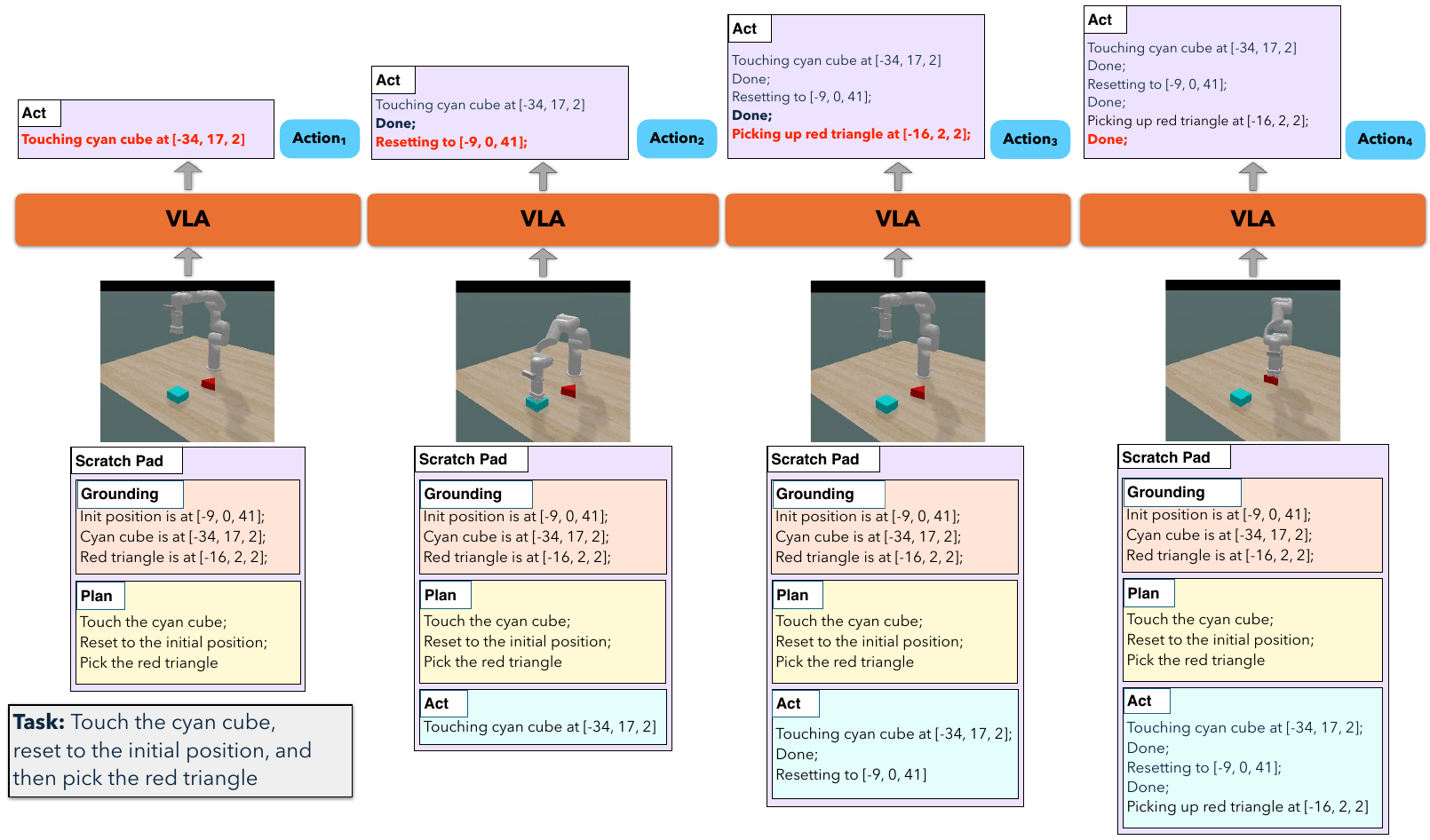}
  \caption{Scratchpad-augmented VLAs. The VLA generates and updates scratchpad which is stored and provided as part of the input context for all subsequent steps, creating an explicit, evolving memory trace of its own behavior.}
  \label{fig:teaser}
\end{figure*}

\begin{abstract}

Many dexterous manipulation tasks are non-markovian in nature, yet little attention has been paid to 
this fact in the recent upsurge of the vision-language-action (VLA) paradigm. 
Although they are successful in bringing internet-scale semantic understanding to robotics, 
existing VLAs are primarily ``stateless'' and struggle with memory-dependent long horizon tasks. 
In this work, we explore a way to impart both spatial and temporal memory to a VLA by incorporating a 
language scratchpad. 
The scratchpad makes it possible to memorize task-specific information, such as object positions, 
and it allows the model to keep track of a plan and progress towards subgoals within that plan. 
We evaluate this approach on a split of memory-dependent 
tasks from the ClevrSkills environment, on MemoryBench, as well as on a challenging real-world pick-and-place task. 
We show that incorporating a language scratchpad significantly improves generalization on these tasks for both non-recurrent and recurrent models. 
\end{abstract}

\section{Introduction}

The advent of LLMs and VLMs has started the paradigm of scaling learning to large data, leading to human-like generalization on a myriad of tasks. Vision-language-action (VLA)~\cite{brohan2022rt, brohan2023rt, kim2024openvla, black2024pi0} models are the instantiation of that paradigm in Robotics. VLAs leverage the rich internet-scale representations learned by VLMs and adapt them to predict actions in the robotics domain. These models are trained end-to-end to take language instructions and camera images as inputs and to predict low-level actions to be performed by the robot. 

VLAs trained on a large amount of robotics data~\cite{o2024open} have shown great promise in generalization to new tasks, robot morphologies, objects and lightning conditions. %
However, despite this progress, the models are still largely ``reactive'' in nature, as they are defined to map a visual input at any given timestep to an action independently of past observations or  actions. 
This makes the policies fail in memory-dependent tasks, where the policy either needs to make sense of the evolving state of the world over a long time-horizon or its own past actions. 

Accordingly, the impressive recent progress in vision-language-action modeling has been confined  
almost exclusively to Markovian tasks, where the current state of the environment suffices 
to predict the next action. 
However, many tasks in the real world require temporal or spatial memory to be solved. 
Consider as a simple example of the task of swapping two objects: 
To swap the positions of the objects, a model first needs to remember their initial positions. 
Then, once it has moved one of them, it needs to remember which object it moved and which one it needs to move next. 
Contemporary VLAs fail on this task, as they are a) unable to remember the object positions, and b) unable to remember which 
sub-task they have already completed and what to do next.

In this work, we investigate how VLAs can be endowed with a memory of their past interactions to succeed in complex, multi-stage tasks. We introduce a simple yet effective strategy of training VLAs with a language scratchpad. The core idea is to leverage the innate language capabilities of the underlying VLM. Before executing an action, the model first predicts a textual description of what it intends to do. This ``thought" is then stored and provided as part of the input context for all subsequent steps, creating an explicit, evolving memory trace of its own behavior. This allows the model to condition future actions not just on the current visual input and the original instruction, but also on a rich, semantic history of its own actions. Since the thoughts generated by VLMs can be very expressive descriptions of the environment, the language scratchpad not only imparts temporal memory but also spatial memory. 

We evaluate our approach on three kinds of memory-dependent tasks. 
First, we introduce ClevrSkills-Mem, a split of memory-dependent tasks from the ClevrSkills~\cite{haresh2024clevrskills} environment aimed at evaluating both temporal and spatial memory capabilities of robotic policies. 
We show that equipping either recurrent or non-recurrent policies with language scratchpad can lead to significant improvements in performance across all these tasks.
Second, we evaluate the approach on MemoryBench~\cite{fang2025sam2act} and show that a scratchpad allows task-agnostic generalist VLAs to achieve performance similar to highly specialized, task-specific methods. 
Third, we show results on a challenging \textit{real-world} memory-dependent pick-and-place task, demonstrating that 
a scratchpad can enable a VLA to perform this task successfully. 
On all tasks, our experiments show significant performance gains with incorporation of a language scratchpad in VLAs. 

In summary, our contributions are as follows:
\begin{itemize}
    \item We present a simple memory mechanism using language based scratchpad, that allows us to train a VLA to create a plan of actions before performing them and to use it to condition all next actions. 
    \item To test this approach, we present a memory dependent split of the ClevrSkills benchmark including five tasks (Touch-Reset-Pick, Place-Next-to-and-Restore, Swap, Stack-and-Topple and Rotate-Restore) where the first two are new tasks, and the last three are taken 
    from~\cite{haresh2024clevrskills}.
    \item We show that both recurrent and non-recurrent policies show significant improvement in performance on memory dependent tasks ($\approx 48\%$ and $\approx$ 11\% for non-recurrent and recurrent VLAs respectively).
\end{itemize}

\section{Related Work}

\paragraph{Vision Language Action models.}
Recent efforts in robotics have focused on creating general-purpose policies by adapting large-scale models. Early approaches involved training large transformer architectures on extensive robotics datasets, pioneered by models like Gato~\cite{reed2022generalist}, which handled a mix of vision, language and robotics tasks, and RT-1~\cite{brohan2022rt}, which focused on robotics data. This trend continued with open-source policies like Octo~\cite{octo_2023} and JAT~\cite{gallouedec2024jack}, which are designed for efficient fine-tuning.

More recently, the trend has shifted to fine-tuning pre-trained Vision-Language Models (VLMs) to leverage their powerful generalization capabilities. This is exemplified by the closed-source RT-2~\cite{brohan2023rt} and similar open-source efforts like OpenVLA~\cite{kim2024openvla} and $\pi_0$ and $\pi_{0.5}$~\cite{black2024pi0, black2025pi05}. 
More recent work has explored various ways to incorporate spatial awareness~\cite{qu2025spatialvla}, 3D perception~\cite{zhen20243d}, streaming and recurrence~\cite{li2023vision, liu2024robomamba, schmied2024large, haresh2024clevrskills}, and embodied chain-of-thought reasoning~\cite{zawalski2024robotic, li2024llara} in these VLAs to improve generalization and efficiency. However, incorporating memory into VLAs has been largely missing from the literature. In this work, we explore a way to incorporate memory in these models to enable them to solve memory-dependent tasks.

\paragraph{Memory in robotics.}
Memory is crucial in enabling robots to achieve complex tasks~\cite{jockel2008towards}. A large portion of the navigation literature has explored incorporating spatial memory by building explicit maps of the environment~\cite{bailey2006simultaneous, chaplot2020object, fuentes2015visual} or attention based memory modules~\cite{mezghan2022memory, fang2019scene}. \cite{wijmans2023emergence} also explored emergence of maps in recurrent navigation architectures. Recent works leveraging vision-language models use 3D representations like voxel-maps, neural-fields and gaussian splats to encode spatial features of the environment~\cite{huang2024copa, huang2023voxposer}. Alternative frameworks have explored video tracking as a mechanism to reason about unobserved, disappearing and reappearing objects in the scene~\cite{huang2024out}. 
Very recenet work SAM2Act~\cite{fang2025sam2act} explores incorporating a memory bank and a memory attention mechanism in a multi-view transformer architecture enabling it to perform tasks requiring spatial memory. In contrast, we leverage the flexibility of language to encode both temporal and spatial context and the inherent capability of VLAs to predict language to endow them with memory. While being very flexible, our framework is also compatible with state-of-the-art reasoning models in both robotics~\cite{zawalski2024robotic,kim2025robot} and vision-language modeling~\cite{nye2021show}.

\paragraph{Recurrent policies in robotics.}

Recent Vision-Language-Action (VLA) models, such as Octo~\cite{octo_2023}, RoboVLMs~\cite{liu2025towards}, and Interleave-VLA~\cite{fan2025interleave}, have successfully processed robotic video data using an interleaved image-text format. A key limitation of these approaches, however, is their reliance on a fixed-size context window, which restricts their memory to only the most recent observations. While recurrent architectures present a natural alternative for maintaining long-term memory, prior works like RoboMamba~\cite{liu2024robomamba}, RoboFlamingo~\cite{li2023vision}, and ClevrSkills~\cite{haresh2024clevrskills} have primarily explored them as an efficient substitute for transformers rather than as a solution for long-horizon memory. Our work builds on efforts to train recurrent policies like XLSTM~\cite{schmied2024large} on large-scale robotics data. We demonstrate that as task horizons lengthen, these recurrent policies still struggle to retain crucial task-relevant information. 
Critically, we find that the performance of recurrent networks can be substantially improved 
by incorporating an explicit separate memory mechanism that aids in information retention over 
temporally extended tasks.
Our findings echo more general observations made in the recurrent network community about the 
necessity of memory-augmentation in tasks that go beyond simple state tracking or 
regular language modeling (see, for example, \cite{deletang2023neural}).

\section{Method}
In this section, we introduce our scratchpad-augmented VLAs and how to generate data to train them. 

\textbf{Preliminaries.} We leverage Vision-Language-Action (VLA) models as our base models. VLAs generally start from a pre-trained VLM and are finetuned on robotics data to produce robot actions auto-regressively. More concretely: given current observation $o_t$, and a language instruction $l$, VLAs model the distribution $p(a_t|o_t, l)$ where $a_t$ is the action at a given timestep. 
Actions are often disctretized into $256$ bins per dimension and mapped to tokens in the VLMs vocabulary~\cite{kim2024openvla}.  

More recent VLAs, however, are also able to reason about tasks in language~\cite{zawalski2024robotic}. More specifically, along with each action, these VLAs are able to predict a language description $d$ of the task and the environment leading to a model $p(a_t, d_t| o_t, l)$. The description can be as simple as stating the current task or as descriptive as including object positions and descriptions of low level actions, such as the direction of the end-effector~\cite{zawalski2024robotic}. 

\subsection{Scratchpad-augmented VLAs.} While they are powerful, current state of the art VLAs~\cite{kim2024openvla, black2024pi0, black2025pi05} are stateless i.e. they look at the current observation and predict the next action without taking any history into account.
This renders them ineffective in memory-dependent tasks, where the policy needs to remember its own actions or the past state of the environment. 
In this work, we propose the use of an external scratchpad of language 
descriptions $S_t=\{d_1, d_2, \dots d_n\}$ as a means to enable an existing VLA to solve such tasks. 
Formally, along with the current observation $o_t$ and the language instruction $l$, 
the policy takes in the scratchpad $S_t$ as the input, and it produces an action and a description $d_t$ as the output, leading to a 
model of the conditional distribution $p(a_t, d_t| o_t, S_t, l)$. 

\begin{algorithm}
    \caption{Scratchpad Update Logic}
    \label{alg:scratchpad_update}
    \begin{algorithmic}[1]
        \Require
            \Statex $p$: The policy.
            \Statex $o_t$: The observation from the environment at timestep $t$.
            \Statex $S_t$: The scratchpad at timestep $t$.
            \Statex $l$: The high-level language instruction or goal.
        \Statex %
        \Function{ForwardPass}{$p, o_t, S_t, l$}
            \State $(a_t, d_t) \gets p(o_t, S_t, l)$ \\ \Comment{Generate action and description}
            \If{$d_t \in d^{\text{update}}$} 
            \\ \Comment{Check if the description triggers an update}
                \State $S_{t+1} \gets S_t \cup d_t$ \\ \Comment{Update scratchpad by adding the description}
            \Else
                \State $S_{t+1} \gets S_t$ 
                \Comment{Keep scratchpad unchanged}
            \EndIf
            \State \Return $(a_t, S_{t+1})$ \\ \Comment{Return the action and the new scratchpad}
        \EndFunction
    \end{algorithmic}
\end{algorithm}

Next, we describe our proposed scratchpad in more detail. The scratchpad is updated by the VLA when necessary, i.e. when a sub-task is successfully completed by the VLA. Specifically, the VLA predicts a special update token that triggers the next description to be included in the scratchpad as seen in Alg.~\ref{alg:scratchpad_update}. 
In practice, the model is trained to recognize the completion of the current sub-task and produce a special $<done>$ token which triggers the description to be updated in the scratchpad. 
Thus, conditioning our proposed scratchpad lends the model a language defined memory. 
Since the descriptions can be a detailed description of the environment state, the scratchpad 
acts a flexible and extendable memory, allowing the model to perform memory dependent tasks. 
An example instantiation of a scratchpad augmented VLA can be seen in Fig.~\ref{fig:teaser}.

\subsection{Generating scratchpad data}

Memory dependent tasks mainly require two types of memory that we shall refer to as 
\emph{Spatial Memory} and \emph{Temporal Memory}, respectively. 
Here, Spatial memory refers to where objects are placed in the environment and how they change positions during the execution of the task, whereas  
Temporal memory refers to what subtasks are being carried out by the robot. 
We generally define the scratchpad to include \textbf{grounding}, \textbf{plan} and \textbf{act} sections. 
The \textbf{grounding} section includes the initialization conditions of the task, which usually include the object positions and the end-effector position (spatial memory), 
the \textbf{plan} section includes the sub-tasks that the policy needs to carry out to complete the task,  
and the \textbf{act} section is used to track current progress by holding the sub-tasks that have been completed by the policy (temporal memory). 
An example instantiation of a scratchpad for a Touch-Reset-Pick task can be seen in Fig.~\ref{fig:teaser}. Note that we design our scratchpad to be flexible, in that, it does not assume a specific type of position coordinates or specific type of semantic action labels. 
The positions can be either 3D coordinates in the environment or 2D pointers in the image space. 
Similarly, action labels can but do not have to carry semantic meaning for the scratchpad to be effective. 

\subsection{Extending scratchpad to Recurrent VLAs}
In addition to an explicit language defined memory via a scratchpad, we can further impart implicit memory to VLAs by using a recurrent backbone which is largely missing from the state-of-the-art literature. Therefore, we also experiment with VLAs with recurrent backbones to highlight the effectiveness of using scratchpad with such models.
As recurrent methods inherently include memory, at first glace, it is not clear if they require an explicit scratchpad.
However, in case of tasks with long time horizons, it becomes increasingly difficult to train recurrent models on 
full-length trajectories. 
Training of recurrent networks on long trajectories is challenging not only due to GPU memory requirements, 
but also due to training instability and vanishing gradients.
To alleviate the first challenge, recurrent networks are commonly trained on sub-sampled 
trajectories with lengths small enough to fit the GPU memory which, however, 
makes it impossible to retain initial environment conditions in long-horizon memory-dependent tasks. 
To adapt scratchpad to recurrent models, we convert our ground truth trajectories into 
interleaved sequences of language instructions $l$, observations $o$, action $a$ and descriptions $d$. 
Concretely, our recurrent models are trained on sequences $Q \in [l,d_1,o_1,a_1,o_2,a_2,\dots,d_k,o_n,a_n]$ all represented in text with a simple next token prediction task.  
We use special $<image>$ tokens to represent observation positions in the sequences and replace 
these with the image embeddings from a vision head after tokenization.

\begin{figure*}[t!]
  \centering
  \includegraphics[width=.95\textwidth,]{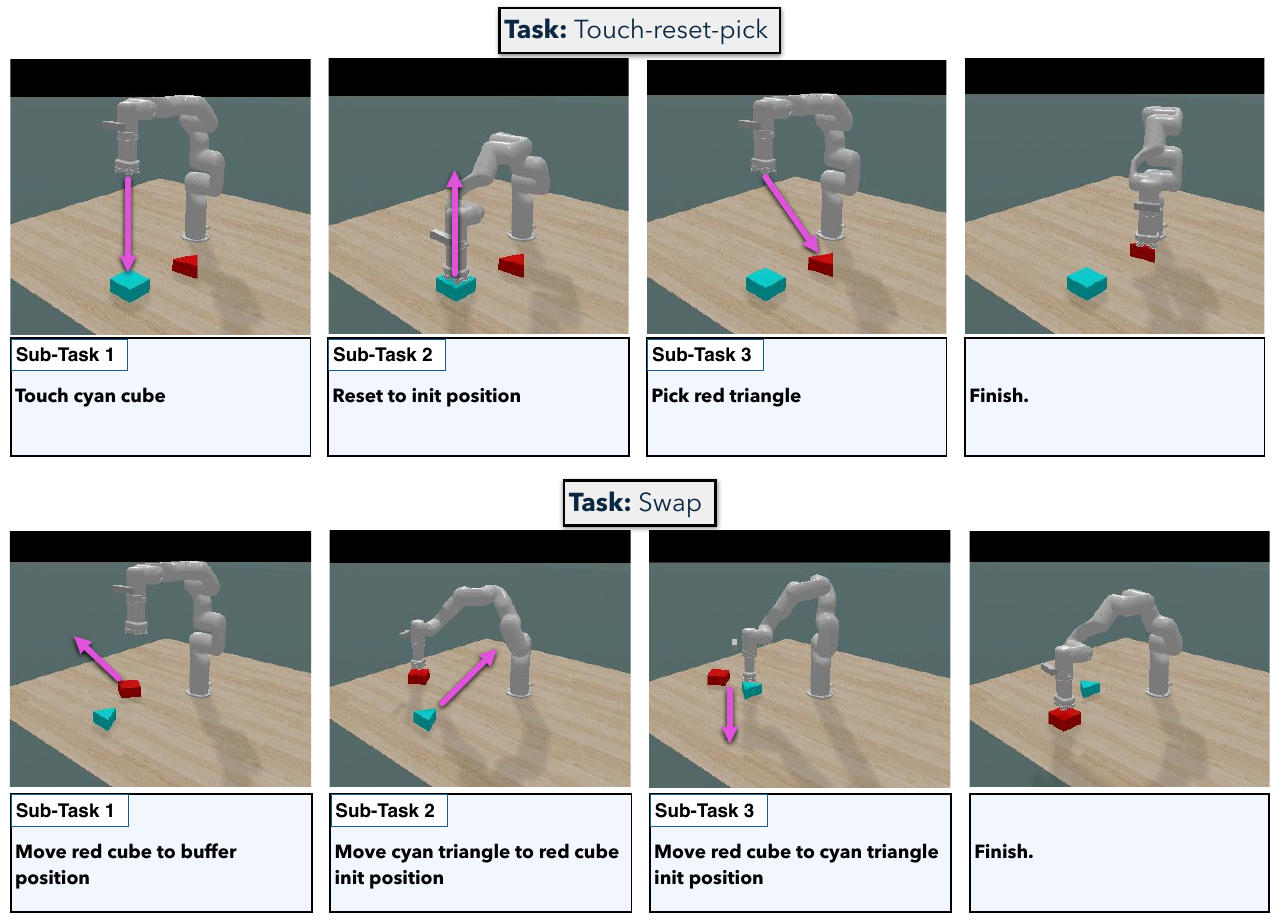}
  \caption{Example memory dependent tasks from ClevrSkills-Mem we evaluate on.}
  \label{fig:tasks}
\end{figure*}

\section{ClevrSkills-Mem Benchmark}

While there is no dearth of robotic simulators and benchmarks, 
memory dependent tasks are largely missing from these. 
Memory dependent tasks violate the Markov assumption prevalent in robotics, according 
to which the current state of the environment and action are sufficient to determine the next state.  
To rigorously evaluate our approach, we leverage the ClevrSkills~\cite{haresh2024clevrskills} benchmark and propose the ClevrSkills-Mem split, which includes tasks that require varying levels 
of information retention to be solved successfully. 
ClevrSkills is based on the ManiSkill2~\cite{gu2023maniskill2} simulator, and it includes oracle solvers to allow generation of ground truth trajectories. 
It adopts a vacuum gripper and uses simple objects, which allows us to focus on the memory aspect of the tasks rather than manipulation complexity of objects. 

ClevrSkills-Mem consists of 5 simple memory-dependent tasks: 
\begin{enumerate}
    \item \textbf{Touch-reset-pick}: the environment is initialized with 2-3 objects and the agent is required to touch a specific object, reset to the initial position and then pick another object. The task requires simple temporal memory, as the agent, once reset, does not know if it needs to do the first sub-task of touching an object or the second of picking an object. 
    \item \textbf{Place-next-to and restore}: the environment is initialized with 2 or more objects and the agent is required to place a specific object near another object. The task evaluates spatial memory, as the agent needs to remember the initial position of the object to be able to restore it to the correct position.  
    \item \textbf{Stack-and-topple}: the environment is initialized with 2-4 objects and the agent is required to stack them in a specific order and then topple the stack. The long-horizon task requires temporal memory to discern which part of the sub-task is being solved at any instance.
    \item \textbf{Swap}: the environment is initialized with 2 or more objects and the agent is required to swap the positions of 2 specific objects. The task evaluates both spatial and temporal memory, as the agent needs to remember initial positions of the objects to be able to swap them correctly, and it furthermore needs to remember which phase of the task it is in to carry out the next sub-task. 
    \item \textbf{Rotate-restore}: the environment is initialized with 2 or more objects, and the agent is required to rotate a specific object by a pre-defined amount and subsequently restore it to the original position. The task requires very fine-grained temporal memory, as once the agent starts rotating the object, it needs to keep track of how much it has rotated it relative to the initial position. 
\end{enumerate}
Fig.~\ref{fig:tasks} shows some examples of these tasks. %

\section{Experiments}

We evaluate scratchpad augmented VLAs in two simulation environments  
(ClevrSkills~\cite{haresh2024clevrskills} and MemoryBench~\cite{fang2025sam2act}) 
and on a real-world 
task using a UFACTORY xArm 6. %

\subsection{Experimental setup}

We use language description for task specification across all models and benchmarks. We resize all images to $224\times224$ before feeding them to the model. 
We use end-effector pose deltas as our action space. 
Actions are 7-dimensional vectors that include 6DOF pose deltas and 1D gripper state.
We experiment with two types of VLA model: 
a transformer based VLA (called \textbf{T-VLA} from here on) which takes a 
single time-slice as input and a recurrent VLA (called \textbf{R-VLA} from here on) which autoregressively generates the whole trajectory.  

For the T-VLA, we linearize the scratchpad and append it with the prompt during training. 
We denote different parts of the scratchpad using special tokens. 
The initialization and plan part of the scratchpad are enclosed in $<plan>...</plan>$ tokens. 
The model is trained to predict the current sub-activity with $<think>$ tokens followed by the action 
where the start is denoted by the $<act>$ token. 
End points of a specific subtask are denoted by the $<done>$ token. 
During evaluation, we let the model predict the $<plan>...</plan>$ on the first observation 
and then follow the algorithm in Alg.~\ref{alg:scratchpad_update} to progressively update 
the scratchpad every time a $<done>$ token is predicted which act as the updated context for 
the next action.

For R-VLA, we interleave the scratchpad and actions in a single long text sequence which follows the format, $[obs_0;plan_0;think_{0};act_{0};obs_1;act_1;...;think_1;act_t...]$. 
Here, observation denotes the image tokens, plan denotes the grounding and plan part of 
the scratchpad, followed by observation, thought and action triples at each timestamp. 
The model is trained on a simple next token prediction task and also evaluated autoregressively in a similar manner. Practically, we found R-VLA to struggle with discretized actions and therefore use a simple MLP as an action head where we represent the action with a $<action>$ token in text and use the embeddings from the action token as input to the action head which then produces the 7-dimensional real valued vectors for actions.

For T-VLA, we start from the PaliGemma-2 VLM model with 3B parameters~\cite{steiner2024paligemma}, which is based on the Gemma-2 LLM~\cite{team2024gemma} and on the SigLIP vision encoder~\cite{zhai2023sigmoid}. We perform full-finetuning of the model with a batch size of $8$ per GPU and a learning rate of $2\times10^{-5}$, using the Adam optimizer~\cite{kingma2014adam}. For R-VLA, we start from Mamba~\cite{gu2023mamba} language model with 130M parameters and use a ViT~\cite{dosovitskiy2020image} as an image backbone. We also experimented with 1B and 3B versions of Mamba model but found little to no performance difference and therefore selected the 130M version as the base language model. We attribute this to the fact that Mamba is not a vision-language model and therefore bigger models do not have better pre-trained visual representations as in other vision-language models. We perform full-finetuning of the model with a batch size of 4 per GPU and a learning rate of $10^{-5}$ using the Adam optimizer~\cite{kingma2014adam}. All experiments were performed using PyTorch~\cite{paszke2019pytorch} on 4 Nvidia A100 GPUs.

\subsection{Results on ClevrSkills-Mem}
\begin{figure*}[thpb]
  \centering
  \includegraphics[width=1.0\textwidth]{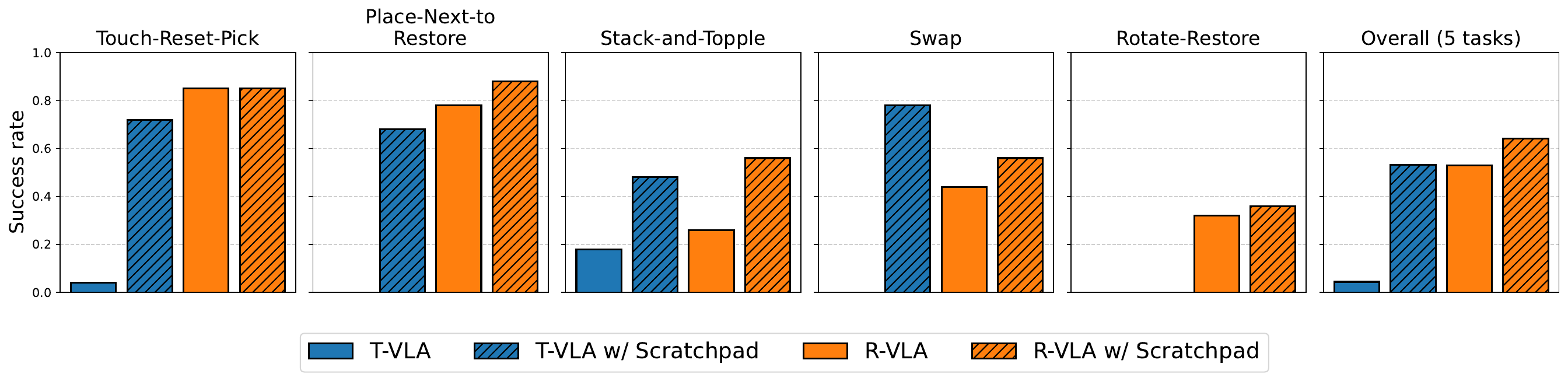}
  \caption{\textbf{Results on the ClevrSkills-Mem benchmark.} We show results for all the models considered with and without scratchpad on 5 tasks of ClevrSkills-Mem benchmark. We report success rate of each model on 50 rollouts on unseen starting positions objects. On the right we show the average performance across all tasks.}
  \label{fig:final_results}
\end{figure*}

The ClevrSkills~\cite{haresh2024clevrskills} benchmark is based on the ManiSkill2~\cite{gu2023maniskill2} manipulation environment, 
and it includes an oracle solver to generate data including actions, object positions, 
and solver traces which are used to get sub-task segmentation to create scratchpad. 
We evaluate the models on a range of tasks that require varying levels of spatial and 
temporal memory. 
Specifically, Touch-reset-pick and Stack-and-topple require strong temporal memory, Swap and Place-next-to-restore require strong spatial memory and Rotate-restore require fine-grained low-level temporal memory. 
Note that tasks requiring spatial memory have a success criterion of $5cm$ delta, i.e. the final object position should not be more than $5cm$ away from the required placement. 
Also note that the objects can be placed anywhere in the workspace and are not located at 
fixed positions as common in other benchmarks. 

We evaluate all models on $50$ rollouts of unseen starting positions of the objects seen during training.
The main results on ClevrSkills can be seen in Fig.~\ref{fig:final_results}. As we can see, due to the challenging nature of the memory-dependent ClevrSkills tasks, T-VLA struggles on all the tasks, and adding the scratchpad significantly improves the performance across all the tasks. 
More specifically, adding the scratchpad improves the performance by 68\%, 72\%, 68\% and 30\% on Touch-reset-pick, Swap, Place-next-to-restore and Stack-and-topple, respectively, leading to 
an average performance gain of 48.8\% across all the tasks. 
We note that T-VLA+Scratchpad still fails to achieve any success on rotate task as it 
requires low-level fine-grained temporal memory. 

\begin{figure}[thpb]
  \centering
  \includegraphics[width=0.95\linewidth,trim={0cm 0cm 0cm 0cm},clip]{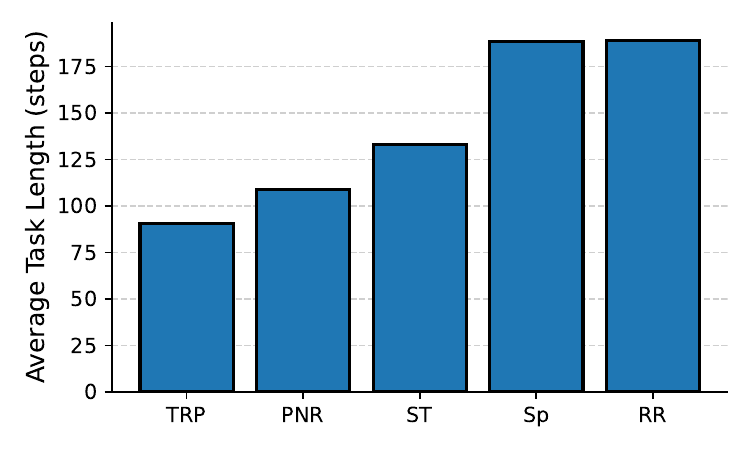}
  \caption{Average trajectory length of tasks in ClevrSkills-Mem. Here, TRP denotes Touch-Reset-Pick, PNR denotes Place-Next-to-Restore, ST denotes Stack-and-Toplle, Sp denotes Swap and RR denotes Rotate-Restore task respectively.}
  \label{fig:task_len}
\end{figure}

We also experiment with a recurrent VLA (R-VLA) with Mamba as backbone to compare against a model with inherent memory. While R-VLA usually performs better than T-VLA, we note that T-VLA+Scratchpad is able to match R-VLA performance on average even surpassing it on some tasks. This shows that language scratchpad is flexible enough to endow strong memory capabilities to stateless VLAs. We also note that recurrent VLAs despite having inherent memory capabilities also see an average improvement of 11\% when trained with scratchpad. In Fig.~\ref{fig:task_len}, we show the average trajectory length of the tasks in ClevrSkills. As we can see, generally increasing task length leads to a higher performance improvement with scratchpad in R-VLAs with exception of Rotate-Restore which see no performance improvement as the task requires low-level fine-grained temporal memory which can not be aided by a scratchpad.
\begin{figure*}[thpb]
  \centering
  \includegraphics[width=1.0\linewidth,trim={0cm 12cm 0cm 0cm},clip]{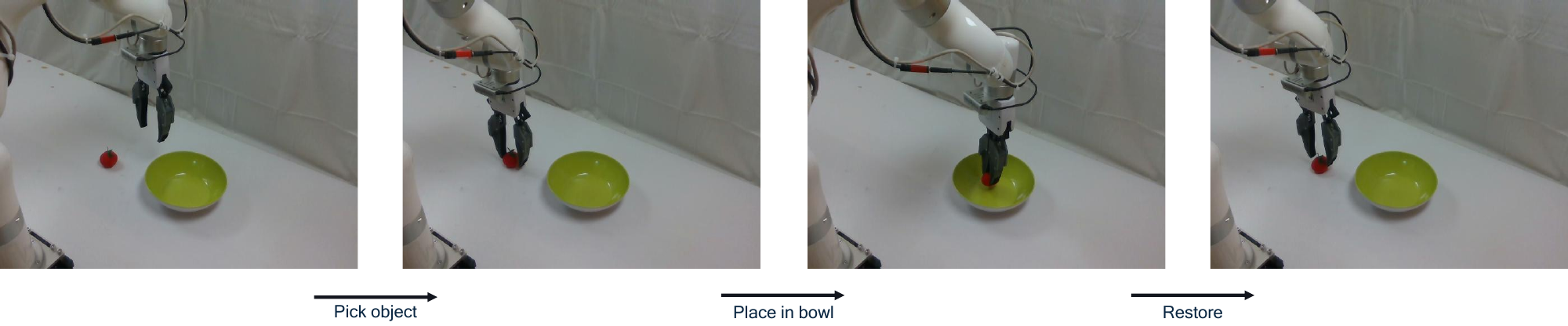}
  \caption{Key frames of the real world task: Pick the tomato, place it in the bowl and then restore it to the initial position.}
  \label{fig:realworld}
\end{figure*}

\subsection{Results on MemoryBench}
To show the flexibility and ability to generalize, we also evaluate the models on the recently proposed MemoryBench~\cite{fang2025sam2act}. 
MemoryBench is built on top of RLBench~\cite{james2019rlbench}, and it includes 3 tasks: 
Put-Block-Back, Rearrange-Block, and Reopen-Drawer. 
MemoryBench differs from ClevrSkills in that objects are placed at fixed position areas and variations are generated based on which area the objects are placed on. 
For example in Put-Block-Back, the cube can start at 1 of the 4 starting positions 
and the task is to move the cube to the middle and then press a button and move the cube back to the initial position. 
This task requires both temporal and spatial memory, as a policy needs to remember if it has pressed the button or not (temporal) and what the initial position of the object was (spatial). 
Note that along with memory, MemoryBench also requires very precise manipulation capabilities, 
especially for pressing the button, as it is very small compared to the scene and therefore very easy to miss. 
We focus on the Put-Block-Back task from MemoryBench as it encompasses both types of memory.

\begin{table}[t]
\caption{Evaluation on Put-Block-Back task from MemoryBench~\cite{fang2025sam2act}.}
\label{tab:membench_results}
\centering
\begin{tabular}{lc}
\hline
\textbf{Models} & \textbf{Put Block Back}\\
\hline
RVT-2~\cite{goyal2024rvt} & $50.0$ \\
SAM2Act~\cite{fang2025sam2act} & $35.0$ \\
SAM2Act+~\cite{fang2025sam2act} & $\textbf{100.0}$ \\
\hline
T-VLA & 0 \\
T-VLA (sim-eval) & 0 \\
T-VLA w/ SP (Our) & 40.0 \\
T-VLA w/ SP (Our) (sim-eval) & \textbf{100.0} \\
\hline
\end{tabular}
\end{table}

To evaluate on MemoryBench, we use the open-sourced 100 trajectories to train our model and evaluate on the 25 trajectories as done in prior work~\cite{fang2025sam2act}. Since MemoryBench requires very precise manipulation, we also use the wrist image as input during training. We follow the prior work to predict the key-frame poses of the end-effector as actions.    

We show performance of our T-VLA and T-VLA+Scratchpad on MemoryBench in Table~\ref{tab:membench_results}. Prior baselines evaluated on MemoryBench all take scene point clouds as input and are driven towards very precise manipulation whereas we just take front RGB and wrist RGB images as input. As we can see, scratchpad significantly improves performance over the stateless baseline. We also note that since we take RGB images as input as opposed to 3D point clouds, our model struggles to press the button and majority of the failure cases stem from this failure mode. Therefore, we also report a relaxed evaluation criteria (denoted by ``sim-eval'' in Table~\ref{tab:membench_results}) where if the VLA gets close to the button, we execute the ground-truth action to press that button instead of taking the action predicted by the VLA and continue the rest of the trajectory. Doing so leads to our model getting a 100\% success rate on the Put-Block-Back task showing that our model is able to perform both temporal and spatial recall perfectly. 

\subsection{Results in the real world}

We also evaluate scratchpad augmented VLAs on a real world task, Pick-Place-Restore, where we task the robot to 
pick an object (a tomato), place it in a bowl and then restore its position to the original position. We show an example trajectory of the task in Fig.~\ref{fig:realworld}. 
Note that we do not assign fixed starting positions for the tomato and the bowl and therefore both 
of them can be located anywhere in the workspace. 
To be successful at the task, the policy needs to remember what sub-task it is currently performing
(temporal memory), as placing the object in the bowl and restoring it to the original position are both ``reverse'' of each other. This leads to a temporal ambiguity when looking at a single 
time-slice at a time.
The policy also needs to remember the starting position of the object in the workspace (spatial memory) for it to be successful at restoring it to the original position. 

We collected a dataset comprising $200$ trajectories from the Pick-Place-Restore task by 
tele-operating a UFactory xArm 6 with a flexible two-fingered gripper, operating on a white tabletop. 
The agent observes the environment through RGB images captured by a RealSense D435 camera, 
positioned at a corner of the table. 
We compare a pretrained OpenVLA model with 7B parameters ~\cite{kim2024openvla} to the 
scratchpad-augmented version of the same model. 
We perform LoRA finetuning with rank 32~\cite{hu2022lora}, 
a batch size of 8 and a learning rate of $5\times10^{-4}$, using the Adam optimizer~\cite{kingma2014adam}.

\begin{table}[t]
\caption{Evaluation on the Real World task.}
\label{tab:real_world_results}
\centering
\begin{tabular}{lc|cc}
\hline
\textbf{Models} & \textbf{Avg. Success} & \textbf{Sub-task CR} & \textbf{Avg. Dist.} \\
\hline
Human & 100\% & 3 & 3.2cm  \\
\hdashline
OpenVLA & 0\% & 0.9 & N/A  \\
OpenVLA w/ SP (Our) & 65\% & 2.4 & 10.62cm \\
\hline
\end{tabular}
\end{table}

We present the results in Table~\ref{tab:real_world_results}. The table shows the average success rate, average sub-task completion rate (denoted by Sub-task CR) and the average distance between the starting position of the object and the final restored position (denoted by Avg. Dist.). 
We evaluate each model on 20 rollouts, where the starting position of the objects are randomized every time. 
As we can see, the VLA without scratchpad fails on this task, as it has no memory of what it is doing. 
Qualitatively, we note that it generally is able to pick the object but restores it back without putting it in the bowl as it can not discern between the start and end of the trajectory as both are roughly identical in the dataset. 
In contrast, the scratchpad augmented model is able to complete all the subtasks $65\%$ of the time demonstrating the memory capacity of the scratchpad. 
We note that the average replacement distance for the model is relatively high as compared 
to the ground truth trajectories. This can be attributed to the limited amount 
of training data (200 examples) given the large variability in the starting positions.

\section{Conclusion}

In this work, we present a simple language scratchpad mechanism that allows us to endow existing VLAs with a flexible and strong memory mechanism enabling them to solve memory-dependent tasks. 
We evaluate the approach in both simulation and in the real world. 
As the number of existing simulation environments for testing long-range memory is limited, we also introduce the 
ClevrSkills-Mem split within the ClevrSkills environment, consisting of existing and novels tasks. 
Together these tasks test for a spectrum of different memory-capabilities with respect to spatial and temporal information. 
Experiments show that scratchpad-augmented VLAs significantly outperform contemporary stateless VLAs for memory-dependent tasks.

\bibliographystyle{./IEEEtran} %
\bibliography{./IEEEabrv,./IEEEexample}

\end{document}